\documentclass[10pt,twocolumn,letterpaper]{article}

\usepackage{iccv}
\usepackage{times}
\usepackage{epsfig}
\usepackage{graphicx}
\usepackage{amsmath}
\usepackage{amssymb}
\usepackage{subcaption}
\usepackage{enumitem}
\usepackage{algpseudocode} 
\usepackage{program}
\usepackage[ruled,vlined]{algorithm2e}

\renewcommand{\vec}[1]{\mathbf{#1}}

\usepackage[pagebackref=true,breaklinks=true,letterpaper=true,colorlinks,bookmarks=false]{hyperref}

\iccvfinalcopy 

\def\iccvPaperID{1823} 

\ificcvfinal\fi

\begin{document}

\title{LMPNet for Weakly-supervised Keypoint Discovery} 

\author{Pei Guo \space Ryan Farrell\\
Brigham Young University\\
{\tt\small peiiguo@gmail.com} \space {\tt\small farrell@cs.byu.edu}
}

\maketitle
\ificcvfinal\thispagestyle{empty}\fi

\begin{abstract}

In this work, we explore the task of semantic object keypoint discovery weakly-supervised by only category labels. This is achieved by transforming discriminatively-trained intermediate layer filters into keypoint detectors.
We begin by identifying three preferred characteristics of keypoint detectors: (i) spatially sparse activations, (ii) consistency and (iii) diversity. 
Instead of relying on hand-crafted loss terms, a novel computationally-efficient leaky max pooling (LMP) layer is proposed to {explicitly} encourage final conv-layer filters to learn ``non-repeatable local patterns'' that are well aligned with object keypoints. 
Informed by visualizations, a simple yet effective selection strategy is proposed to ensure consistent filter activations and attention mask-out is then applied to force the network to distribute its attention to the whole object instead of just the most discriminative region. For the final keypoint prediction, a learnable clustering layer is proposed to group {keypoint proposals} into keypoint predictions.
The final model, named LMPNet, is highly interpretable in that it directly manipulates network filters to detect predefined concepts. Our experiments show that LMPNet can (i) automatically discover semantic keypoints that are robust to object pose and (ii) achieves strong prediction accuracy comparable to a supervised pose estimation model.

\end{abstract}

%
%

\section{Introduction}\label{sec:intro}
Although according to Aristotle, ``the whole is greater than the sum of its parts'', representing an object as a collection of features from its part constellation is proven to be more powerful~\cite{1335452,felzenszwalb2009object,zhang2014part} than holistic object-level representations. Part-based representations have been widely adopted by tasks like classical object detection~\cite{1335452,felzenszwalb2009object}, fine-grained visual categorization~\cite{berg2013poof,zhang2014part}, and person re-identification~\cite{li2017learning}, \textit{etc.} As an indispensable component in the part-based representation, object keypoint localization has been well studied in previous works~\cite{Cao_2017_CVPR, 10.1007/978-3-319-46484-8_29,6909610}. Different from supervised algorithms whose reliance on manual annotation may hinder their application to large-scale datasets, 
we pursue a different route for keypoint discovery, weakly-supervised by easier-to-obtain object category labels. Beside its obvious benefits of reduced annotation cost, weakly-supersived keypoint discovery has its own virtue that enables us to manipulate and transform discriminatively-trained filters to learn  semantic features as desired using carefully-designed network components. 

\begin{figure}[t]
\begin{center}
\includegraphics[width=\linewidth]{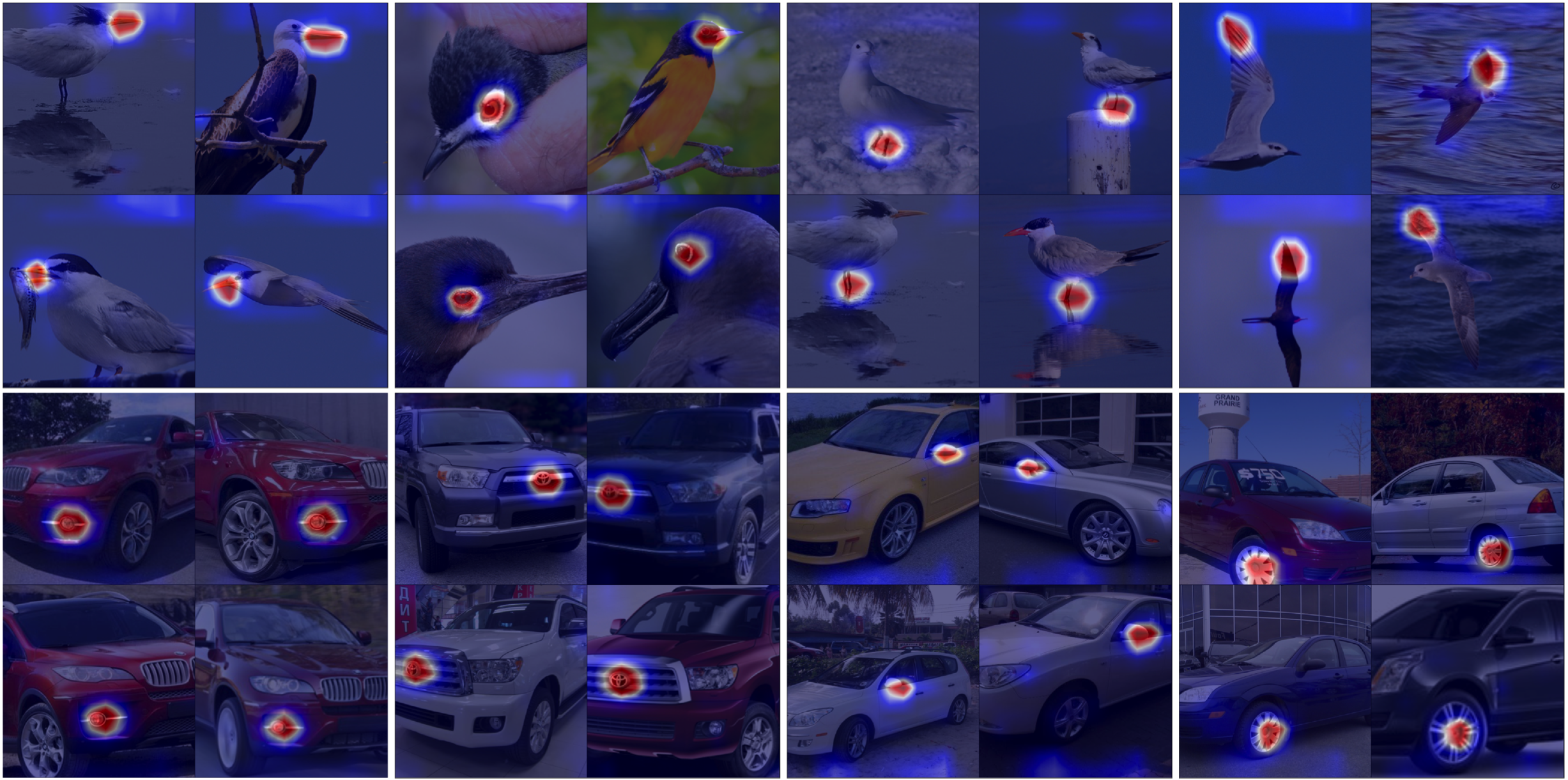}
\caption{{\textbf{Final convolutional layer filter visualization.} The proposed leaky max pooling layer helps the network learn ``non-repeatable local patterns'' that align well with object keypoints. Each $2\times2$ region represents one final-conv layer filter with activations. The detected object keypoints include bird beak, eyes, feet and wing tip; vehicle logo, rear-view camera, wheels and frontal light.}}
\label{fig:headline}
\end{center}
\end{figure}

\begin{figure*}[ht]
  \centering
  \includegraphics[width=\linewidth]{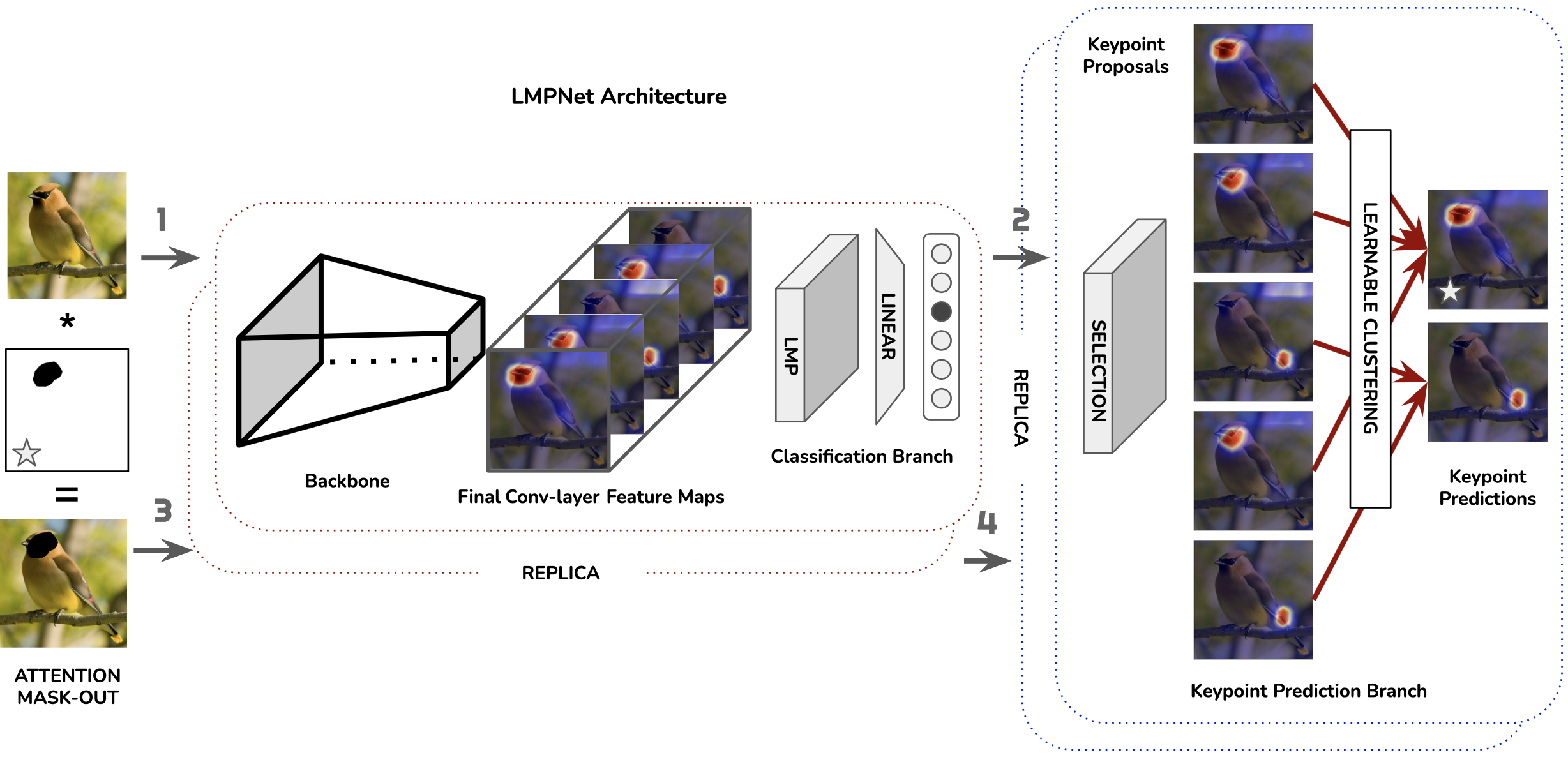}
  \caption{\textbf{LMPNet Architecture.} The final conv-layer feature maps are fed into two branches: a classification branch with leaky max pooling and a keypoint prediction branch with selection and learnable clustering. The most discriminative region (the first keypoint prediction) is then masked out from the input image and fed into a replica network.}
  \label{fig:big_pic}
\end{figure*}

Weakly-supervised learning~\cite{ahn2019weakly,Choe_2019_CVPR,hwang2021weakly,novotny2017anchornet,song2014learning,yao2020saliency,zhang2018unsupervised,Zhou_2016_CVPR,zhou2018weakly} has gained popularity in the research community over the last few years. Representative tasks include object detection~\cite{Choe_2019_CVPR,song2014learning,Zhou_2016_CVPR}, semantic segmentation~\cite{yao2020saliency,ahn2019weakly}, instance segmentation~\cite{hwang2021weakly,zhou2018weakly} and dense correspondence~\cite{novotny2017anchornet,zhang2018unsupervised} using images labels or captions. Among the extensive research efforts, one particular interesting direction is to leverage the aggregated discriminatively-trained intermediate network filters. We argue that the key to success for filter aggregation is to identify the desired characteristics of the target output and transform the filters accordingly. Although network filters have been shown to learn to detect different concepts during discriminative training~\cite{bau2017network,zhou2014object, Zhou_2016_CVPR}, there is a lack of a systematic ways to manipulate the learned internal representations to be driven by specific needs. Previous algorithms ``guide'' the filter learning process by injecting hand-crafted loss terms~\cite{novotny2017anchornet,zhang2018unsupervised}. However, these losses are only indirectly influencing the network and their relative weights are hyper-parameters that are tricky to balance. In this paper, we propose several novel network components that impose direct constraints on internal network filters and feature maps based on solid evidence that is from neural network interpretation, and our efforts, in return, contributes to a deeper network understanding.

Work from Choe,~\etal~\cite{choe2020evaluating} suggested that weakly-supervised object localization can be ill-posed, as network filters can learn to represent both objects and background, and the aggregation of filter activations is not guaranteed to be within the object boundary. We further argue that weakly-supervised keypoint discovery can be in vain without proper definition of object keypoints. In this paper, a semantic object keypoint is defined as a \emph{non-repeatable local pattern}. Examples of semantic keypoints include: landmarks on the human face (eyes, nose, mouth, \textit{etc.}), the eyes, beaks and feet of a bird, or the logo and headlights of a vehicle. Keypoint-level features are spatially smaller than part-level features, examples of which are the cheek, 
forehead, wings and belly of a bird, and the windshiled and doors of a vehicle. Background regions  (sky, lake, trees, \textit{etc.}) and object-level properties (body color, the pose of flying or perching) are not regarded as keypoints. 

Next, we show how the definition of ``{non-repeatable local patterns}''  leads naturally to the development of a novel global pooling~\cite{lin2013network} layer. Under this definition, keypoint detectors should produce spatially sparse feature maps similar to a narrow spatial Gaussian distribution.  
One key insight to draw is the influence of the global pooling layer on final conv-layer feature map sparsity. Max pooling has been shown to produce sparser feature maps compared to its average pooling counterpart, but the max operation is not optimal as it doesn't prevent a pattern existing everywhere, violating the ``non-repeatable'' rule. An ideal pooling layer should detect a local pattern that \emph{only} exists at the maximum activation location. We thus propose a novel pooling layer, named leaky max pooling, that not only allows one pixel out like max pooling, but also suppresses other pixels with negative weights. Leaky max pooling can be implemented by reformulating the pooling operation as a matrix-vector multiplication, for which average and max pooling are shown to be special cases. Only a tiny amount of computation is introduced between both the forward and the backward processes. Leaky max pooling explicitly encourages sparse activations that are well-aligned with object keypoints. It is a plug-and-play module that can be applied to most modern network architectures.

It can be tempting to assume that filters develop into ``perfect'' keypoint detectors with leaky max pooling, however, two phenomena were observed that challenge this assumption. First, filters do not necessarily consistently activate on a tight cluster of patterns. A significant number of filters actually activate on multiple patterns (a mechanism known as distributed representation). Blindly aggregating filters would lead to poor keypoint predictions. After extensive investigation of the filter visualizations, we find a strong correlation between a filter's activation strength and its selectivity. A higher activation more likely indicates the existence of the certain underlying patterns. A simple yet effective selection strategy is therefore proposed that only keeps highly-selective filters. Second, keypoint proposals are not evenly distributed across the object. A majority of filters end up detecting the same concept -- the most discriminative feature -- 
like the head of a bird. This causes difficulty for the keypoint prediction as keypoints from less discriminative parts are poorly represented in the keypoint proposals.
A key observation is that if the most important image region is masked out, the network will be forced to diversity its attention to other parts.

The feature maps before the learnable clustering layer are named ``keypoint proposals'', which potentially contain object keypoints.
To produce final keypoint predictions from the keypoint proposals, a many-to-one projection is learned where predictions are decided by votes from multiple proposals. The weight between a keypoint proposal and a keypoint prediction gets updated iteratively to reflect their association relationship: keypoint proposals closer to a prediction are assigned larger weights. A greedy strategy is applied to sequentially output keypoint predictions. In each iteration, the most prominent keypoint prediction is produced and proposals that vote for it are erased before the next iteration.

The main contributions of this paper are:

\setlist{nolistsep}
\noindent
\begin{itemize}[noitemsep,leftmargin=*]
    \item Identifying three preferred characteristics for keypoint detectors that are missing in discriminatively-trained filters via network visualization.
    \item Proposing several novel network components that are highly integratable and computationally efficient, including leaky max pooling and learnable clustering, that detect and cluster ``non-repeatable local patterns''.
    \item The proposed network architecture, LMPNet, produces diverse and meaningful object keypoints across different classes, and poses, achieving accuracies comparable to supervised keypoint prediction models.
\end{itemize}


%
%

\section{Related Work}\label{sec:related-work}

Lying between supervised~\cite{NIPS2012_c399862d, HeZRS_CVPR2016} and unsupervised learning~\cite{bengio2012deep,erhan2010does}, weakly-supervised learning~\cite{oquab2015object,zhou2018weakly,novotny2017anchornet} presents unique challenges to the research community. Although supervised learning remains the dominating scheme for major vision tasks, its demand for human labels often tampers its application to the enormous amount of unlabelled datasets. Unsupervised learning automatically learns features from raw image data, but its training goal is tricky to define.  Evaluation of unsupervised learning relies on the definition of good feature representations~\cite{bengio2013representation} like disentanglement, but whether it can be learned without inductive bias or is fundamentally necessary are still under debate~\cite{locatello2019challenging,mathieu2018disentangling,higgins2018towards}. Recently, self-supervised learning~\cite{kolesnikov2019revisiting,jing2020self,noroozi2016unsupervised,gidaris2018unsupervised} gains momentum as it generates training signals from the image itself, by predicting patches' relative location~\cite{noroozi2016unsupervised} or the rotation angle~\cite{gidaris2018unsupervised}, \textit{etc.} The dilemma of unsupervised/self-supervised learning is that they are often used to pretrain a model and strong labels are still needed to for downstream tasks like classification~\cite{noroozi2016unsupervised,gidaris2018unsupervised} and landmark prediction~\cite{jakab2018unsupervised,thewlis2017unsupervised}. 


\textbf{Weakly-supervised object detection (WSOD)} WSOD aims to predict the object bounding box using only image labels. One of the popular models in WSOD is multiple instance learning (MIL)~\cite{maron1998framework}. 
In MIL, a positive bag contains at least one positive sample and a negative bag contains all negative samples. The standard MIL pipeline includes 
the repeating process of detector training and object relocating. In recent years, inspired by network visualization, many algorithms~\cite{Zhou_2016_CVPR,Choe_2019_CVPR,zhou2018weakly,novotny2017anchornet} has been proposed to aggregate the feature maps of a discriminatively-trained CNN model to get the object location. Class activation map (CAM)~\cite{Zhou_2016_CVPR} works by reweighting final conv-layer feature maps using the linear layer weights for WSOD. Choe~\etal~\cite{choe2020evaluating} conducted an extensive evaluation of  WSOD algorithms and found that WSOD can be a ill-posed task with only image labels as supervision. As the only constraint on network filters is the cross entropy loss for classification, they tend to focus on the most discriminative part, or learn background features instead of object features. Aggregation of these filters may lead to incomplete or over-complete bounding boxes.

\textbf{Weakly-supervised part discovery} Early works in fine-grained recognition~\cite{KrauseJYF_CVPR2015,XiaoXYZPZ_CVPR2015,ZhangXZLT_CVPR2016} explores object part detection free of part or keypoint annotation. 
Krause~\etal~\cite{KrauseJYF_CVPR2015} proposed to generate object mask
by co-segmentation and find object parts by aligning the masks. Xiao~\etal~\cite{XiaoXYZPZ_CVPR2015} performs spectral clustering on the intermediate filters to produce filter groups as part detectors. Zhang~\etal~\cite{ZhangXZLT_CVPR2016} instead picks filters that generates top responses as part candidates and trains them by iterative negative sample mining.  Simon and Rodner~\cite{SimonR_ICCV2015} proposed an EM-like optimization process to generate a star shape model for a subset of selected part proposals from the internal representation of CNNs. Compared to our end-to-end trainable LMPNet, the above methods usually involve offline process of filter aggregation and selection.

\textbf{Weakly-/self-supervised landmark detection}
Perhaps the most similar works to ours are AnchorNet~\cite{novotny2017anchornet} and~\cite{zhang2018unsupervised}. A set of filter of diversity, discriminality, and sparsity constraint are learned by the proposed loss terms. \textit{etc.}  A series of works utilize the equivariance constraint of 2D image transformation (translation, rotation, scaling) to learn landmarks automatically, which is similar to the concept of self-supervised learning. To evaluate the generated landmarks, a weak network (three layer MLP) is usually used to regress to the ground truth keypoints. Different from the above methods, our algorithm employs no extra loss terms but proposes novel network components that explicitly transform network filters into keypoint detectors. More over, we employ greedy matching PCK to evaluate our keypoint prediction, without supervised training.


%
%

\begin{figure}[t]
  \centering
  \includegraphics[width=\linewidth]{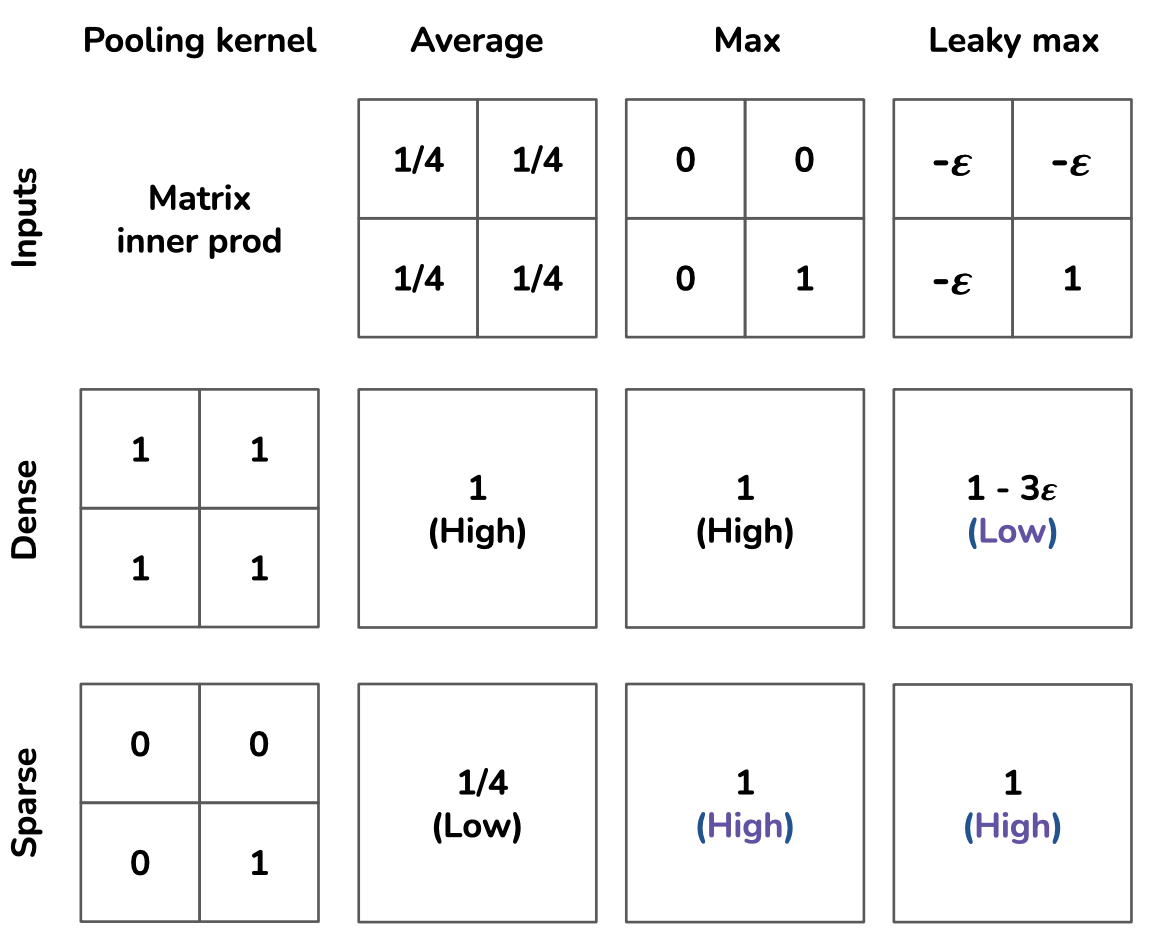}
  \caption{{\textbf{Examples of different pooling methods.} We show activations for dense and sparse input using average, max and leaky max pooling separately. Leaky max pooling produces low output for dense inputs and high output for sparse inputs.}}
  \label{fig:kernel}
\end{figure}

\section{LMPnet Architecture}\label{sec:lmp-net}
A comphrehensive overview of the proposed LMPNet architecture can be found in Figure~\ref{fig:big_pic}. The LMPNet builds on a classification network, replacing the global pooling layer with leaky max pooling. A keypoint prediction branch is added that takes the final conv-layer feature maps as input and produces keypoint predictions. In the keypoint prediction branch, input filters are selected and fed into the learnable clustering layer. 
The original image with the most discriminative region erased is then fed into a replica network that shares the same architecture but with independent weights. The keypoint predictions from the original network and the replica network are fused as the final predictions. The following sections contain detailed descriptions of the proposed LMPNet network modules.


\subsection{Leaky Max Pooling}

We motivate the leaky max pooling layer by the following toy example (Figure~\ref{fig:lmp} (a)). 
Suppose there are two $2\times2$ feature maps: one feature map is densely activated containing all $1$s; the other one is sparsely activated containing all $0$s but one $1$. 
An ideal pooling layer should produce high output for sparse feature maps and low output for dense feature maps.
Let's first examine the two most common pooling types: average and max.
The average pooling outputs $1/4$ for the sparse input and $1$ for the dense inputs. This is undesirable as average pooling encourages dense inputs rather than sparse inputs.
Max pooling outputs $1$ for both sparse and dense inputs. Compared to average pooling, max pooling produces higher output for sparse inputs, but does not discourage dense inputs.
The proposed leaky max pooling produces $1-3\epsilon$ for the dense input, lower than both average and max pooling, and $1$ for the sparse input, the same with max but higher than average.
Therefore, leaky max pooling encourages sparse inputs and suppresses dense inputs at the same time.

Next we show how the global pooling layer can be implemented by a general matrix-vector multiplication operation. 
Suppose the input to the pooling layer is $\vec{X} \in \mathbb{R}^{b\times c\times h \times w}$ where $b,c,h,w$ are batch size, channel number, feature map height and width separately.  The output of pooling layer is $\vec{Y} \in \mathbb{R}^{b \times c}$. The input can be reshaped as a 2D matrix for convenience: $\tilde{\vec{X}} \in \mathbb{R}^{bc \times hw}$. A pooling vector is defined as $\vec{w} \in \mathbb{R}^{hw}$. The pooling operation can then be reformulated as matrix-vector multiplication: $\tilde{\vec{Y}} = \tilde{\vec{X}}\vec{w}$, and $\tilde{\vec{Y}} \in \mathbb{R}^{bc \times 1}$ is then reshaped to $b \times c$. Different pooling operations can be distinguished by their respective pooling vector $\vec{w}$ under such formulation.
It is obvious that for average pooling, \\
\begin{equation}
    \vec{w}_{i} = \frac{1}{hw},\quad \forall i,
\end{equation}
and for max pooling:
\begin{equation}
    \vec{w}_{i} = 
    \begin{cases}
    1 & \text{if } {i} = \text{argmax}(\vec{x})\\ 
    0 & \text{otherwise.} 
    \end{cases}
\end{equation}
The leaky max pooling is defined by:
\begin{equation}
    \vec{w}_{i} = 
    \begin{cases}
    1 & \text{if } {i} = \text{argmax}(\vec{x})\\ 
    -\epsilon & \text{otherwise.} 
    \end{cases}
\end{equation}
where $\epsilon$ is a small positive number. The above operation introduces only a tiny amount of additional computation. Take ResNet-50 for example, the computation overhead (measured in FLOPS) introduced by the matrix-vector multiplication adds only $0.05\%$ to the whole network.

\begin{figure*}[t]
    \centering
    \includegraphics[width=\linewidth]{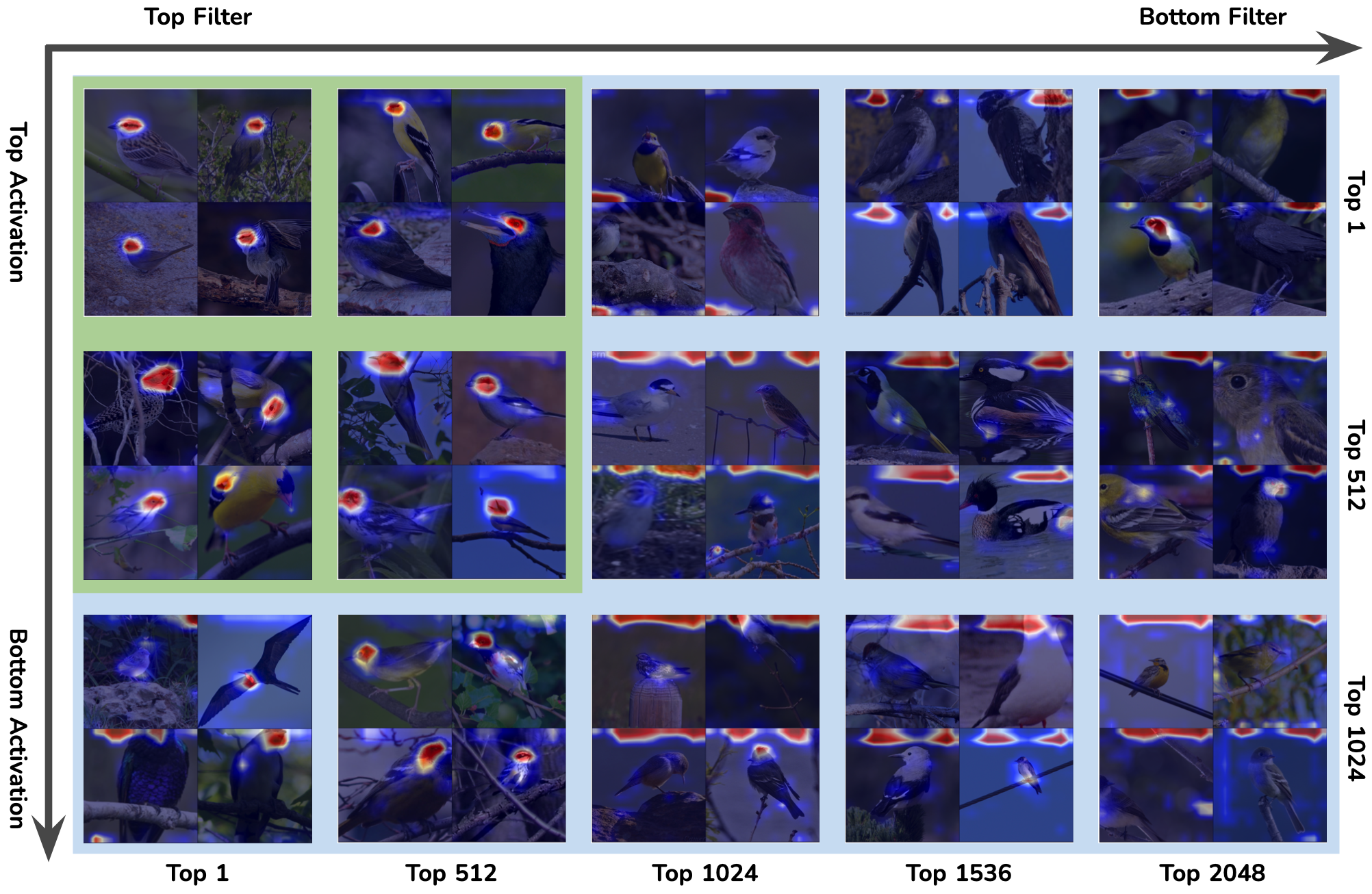}
    \caption{\textbf{Network filter visualization.} Each column represent a filter. Each row represents a group of feature maps that activate a filter. Activation strength decreases from top to bottom and from left to right. We select filters by their activation strength to eliminate noisy filters.}
    \label{fig:full-vis}
\end{figure*}

The gradient distribution of the final conv-layer feature maps provides an even better perspective of leaky max pooling's advantage. Each element inside the feature map  gets a gradient signal proportional to the corresponding weight of the pooling kernel. For average pooling, every feature map pixel gets the same gradient, positively encouraging patterns to exist across the whole image. For max pooling, only the maximum pixel gets non-zero gradient, encouraging a pattern to exist in a local neighborhood. However, all other pixels get zero gradient, meaning they are neither encouraged nor discouraged. They tend to remain unchanged during the subsequent training iterations. This, a pattern that universally exists, has a high probability of being learned by max pooling. The proposed leaky max pooling overcomes the limits of both the average and max pooling methods by promoting only ``non-repeatable local patterns''. Compared to max pooling, non-maximum elements get a negative gradient and are suppressed in the next iteration.
The leaky max pooling kernel can also be thought of as a template that matches only sparsely-activated feature maps.

The final conv-layer feature maps before the leaky max pooling layer are spatially sparse and well-aligned with ground truth keypoints, but it is problematic to feed them directly to the learnable clustering layer without proper post processing. These feature maps are noisy, as a filter does not necessarily activates on one specific pattern.
Moreover, high-level filters often concentrate on the most discriminative region, leading to   unevenly distributed keypoint proposals. In the next two sections, we discuss these issues in detail and describe our methods to tackle them.

\subsection{Filter Consistency}
Network interpretation~\cite{bau2017network,zhou2014object, Zhou_2016_CVPR} via visualization has been well studied in discriminative training. 
Bau~\etal~\cite{bau2017network} further conducted statistical studies of the interpretability of discriminatively-trained network filters and found that not all units are interpretable. For example, only {71\%} of filters are deemed ``interpretable'' in AlexNet. They also found that a network's performance seems to be independent of its filters' interpretability. In fact, Hinton~\cite{hinton1984distributed} argued that the distributed representation, where each neuron is responsible for multiple concepts, is what makes a neural network representation powerful. In our case, a large portion of filters only learn random/noisy features; these filters are the ones with smaller maximum activation strength. This lead us to believe that the network is easily content with its filters as long as they're discriminative. The least-selective filters learn slower and never get a strong enough gradient to update themselves, winding up with activations on random features. 

Another side of the story is that even filters that activate on one specific pattern are not consistent. 
Our network visualization method works by ranking the output feature maps by their max value. Compared to optimization methods,  visualization via dataset samples provides a more realistic and diverse view of a filter's learned pattern; only samples with top activations are typically shown due to space limitations. However, the whole picture of a filter's learned pattern can only be revealed by displaying less-activated feature maps and images.  

To facilitate a comprehensive understanding of the filter-concept relationship, we design a ``full-scale'' filter visualization graph shown in Figure~\ref{fig:full-vis}.  $2\times2$ image tiles are displayed in a 2D grid. Each column represents a filter. Tiles of images in each column are ranked by their images' activation strength, descending from top to bottom. From left to right, filters are ranked by decreasing maximum activation strength. We sample the filters and images/feature maps evenly along each dimension. It can be seen from Figure~\ref{fig:full-vis} that images with higher activations are more likely to contain the underlying pattern (the top left portion of the graph). This suggests that a filter's selectivity (its ability to focus on one specific pattern) is positively related to its activation strength. Intuitively, a higher activation indicates the filter is more confident of the existence of a pattern. We thus propose to use activation strength as a strong clue for filter selection. We simply keep the filters with the highest activations for each input image.

\begin{algorithm}[t]
\SetAlgoLined
\SetKwInOut{Input}{Input}
\SetKwInOut{Output}{Output}
\SetKwInOut{Parameter}{Parameter}
\Input{$\vec{X} \in \mathbb{R}^{c \times h \times w}$\;}
\Output{$\vec{Y} \in \mathbb{R}^{k\times h \times w}$\;}
\Parameter{$\vec{W} \in \mathbb{R}^{c \times k}$\;}
$\vec{W} \leftarrow \vec{1}$\;
 \For{$i\leftarrow 1$ \KwTo $k$}{
    \For{$it\leftarrow 1$ \KwTo $n$}{
        $\vec{w} \leftarrow \text{softmax}(\vec{W}[:,i])$\;
        $Y \leftarrow \vec{w}^T \vec{X}$\;
        $\vec{d} \leftarrow \text{dist}(Y, \vec{X})$\;
        $\vec{W}[:,i] \leftarrow \vec{w} + 1 / \vec{d}$\;
    }
    $\vec{w} \leftarrow \text{softmax}(\vec{W}[:,i])$\;
    $Y \leftarrow \vec{w}^T \vec{X}$\;
    $\vec{Y}[i] \leftarrow Y$\;
    $\vec{X}[\vec{d} < thr] \leftarrow 0$\;
    \If {$max(\vec{X}) = 0$} {break\;}
 }
 \caption{\textbf{Learnable Clustering.}}
\end{algorithm}


\subsection{Filter Diversity}

Unsupervised clustering works best on balanced data where each cluster contains a similar number of data samples. However, filter visualization suggests that discriminatively-trained filters tend to focus on the most informative image region. We propose to tackle this issue using attention mask-out. Mask-out can be used for important image region attribution: tiny image regions are sequentially masked out before the image is passed into the network and the classification result is observed. The regions that cause incorrect classification results are regarded as the attention regions of the network. Our idea is that when the most important region is masked out from the input, the network has to learn to distinguish between categories using other regions, thus distributing its attention to other parts of the object. 

Our attention mask-out strategy consists of two LMPNet with the same architecture, but with independent weights. The first LMPNet takes the original input and outputs initial keypoint proposals. A binary mask is generated by reversing the first proposed keypoint.
The binary mask is multiplied with the original image to remove its most discriminative region. The masked image is then fed into the second, replica LMPNet while training. During the testing stage, the outputs of both networks are fused to get the final keypoint predictions.

\subsection{Learnable Clustering}

We propose a learnable clustering layer to aggregate the keypoint proposals, after selection and attention mask-out, into keypoint predictions. We describe our learnable clustering layer in Algorithm 1. Without loss of generality, we consider the feature map matrix for a single input $\vec{X} \in \mathbb{R}^{c \times h \times w}$, where $c,h,w$ are channel number and feature map height and width. We set the maximum value of each feature map to 1 and everything else to 0 to remove noise.
The layer output is $\vec{Y} \in \mathbb{R}^{k \times h \times w}$, where $k$ is the number of keypoint predictions. 
We define a weight matrix $\vec{W} \in \mathbb{R}^{c \times k}$. The weight matrix elements are initialized to 1. The forward procedure includes $n$ iterations. In each iteration the keypoint predictions are generated by weighting over keypoint proposals using $\vec{W}$. The distance between keypoint proposals $\vec{X}$ and keypoint prediction $Y$ is computed using dist() as the distance between their max value positions. The similarity between $\vec{X}$ and $Y$ is the reciprocal of their distance.
$\vec{W}$ is updated to reflect the accumulated similarity between the input and output and to account for the contributions of each keypoint proposal to the keypoint prediction. Keypoint proposals that are closer the prediction get larger weights and contribute more to the final result. An erasing step $\vec{X}[\vec{d} < thr] \leftarrow 0$ is performed after each keypoint prediction so that keypoint proposals that belong to previous predictions have no influence on new predictions. This is a similar idea to non-maximum suppression~\cite{canny1986computational,rothe2014non} in object detection, where object proposals with the highest score are sequentially selected and   overlapping proposals are removed. These iterations are repeated until no proposals remain or until all $k$ keypoints are found.

\begin{figure}[t]
  \centering
  \includegraphics[width=\linewidth]{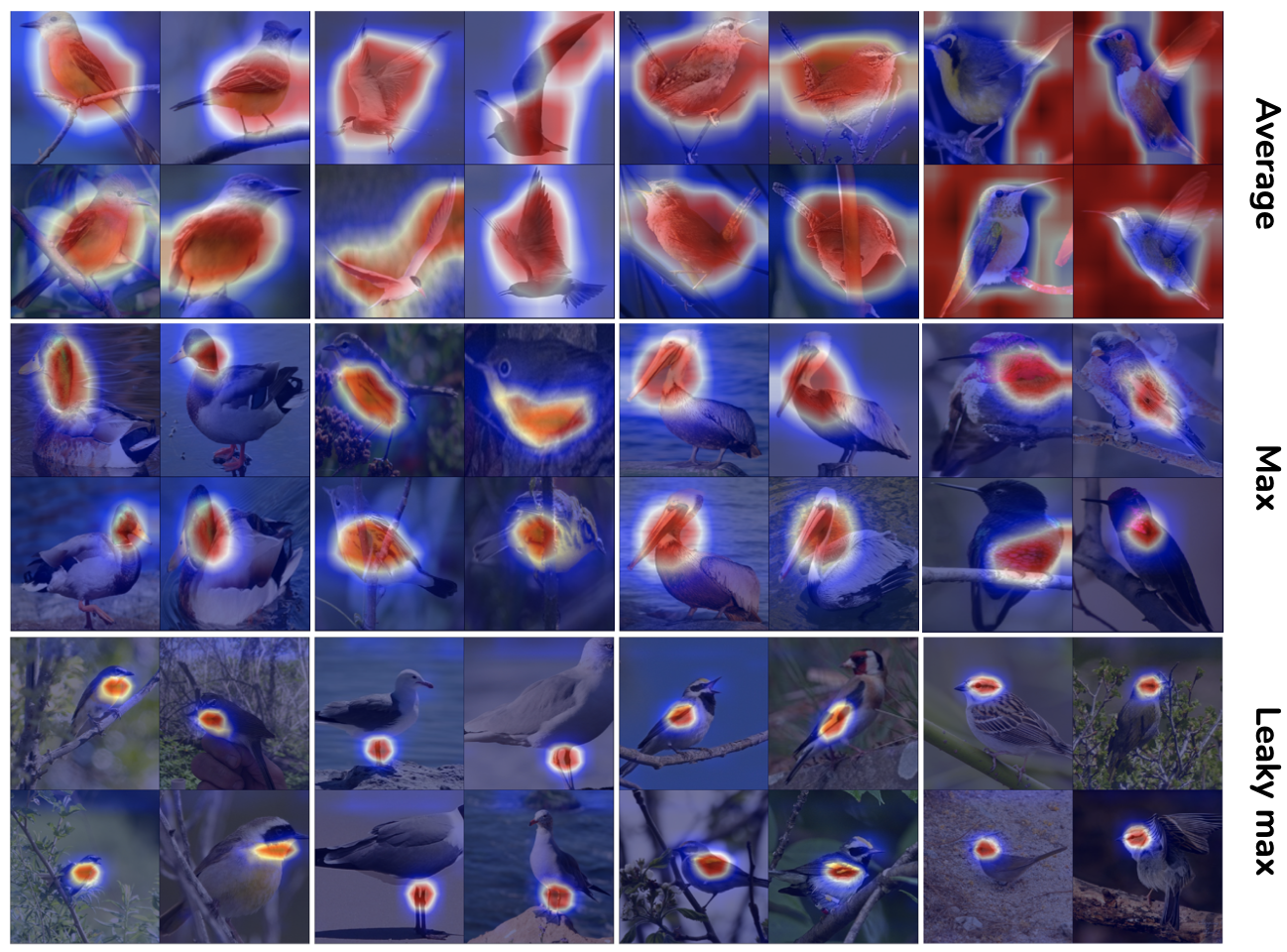}
  \caption{{\textbf{Feature map sparsity comparison.} From top to bottom are filter visualizations obtained using average, max and leaky max pooling. Average pooling learns object-level and background features, max pooling encourages part-level features, and leaky-max pooling focuses on keypoint-level features. See appendix for more visualization results.}}
  \label{fig:comparison}
\end{figure}

\section{Experiment} \label{sec:exper-analysis}
This section contains analysis of the leaky max pooling layer and evaluation of the keypoint prediction results.  
The fine-grained classification dataset CUB-200-2011~\cite{WahBWPB_Tech2011} is used in these experiments. ResNet-50~\cite{HeZRS_CVPR2016} is used as the backbone model.

\subsection{Leaky Max Pooling Analysis}
\begin{figure}[t]
    \centering
    \includegraphics[width=\linewidth]{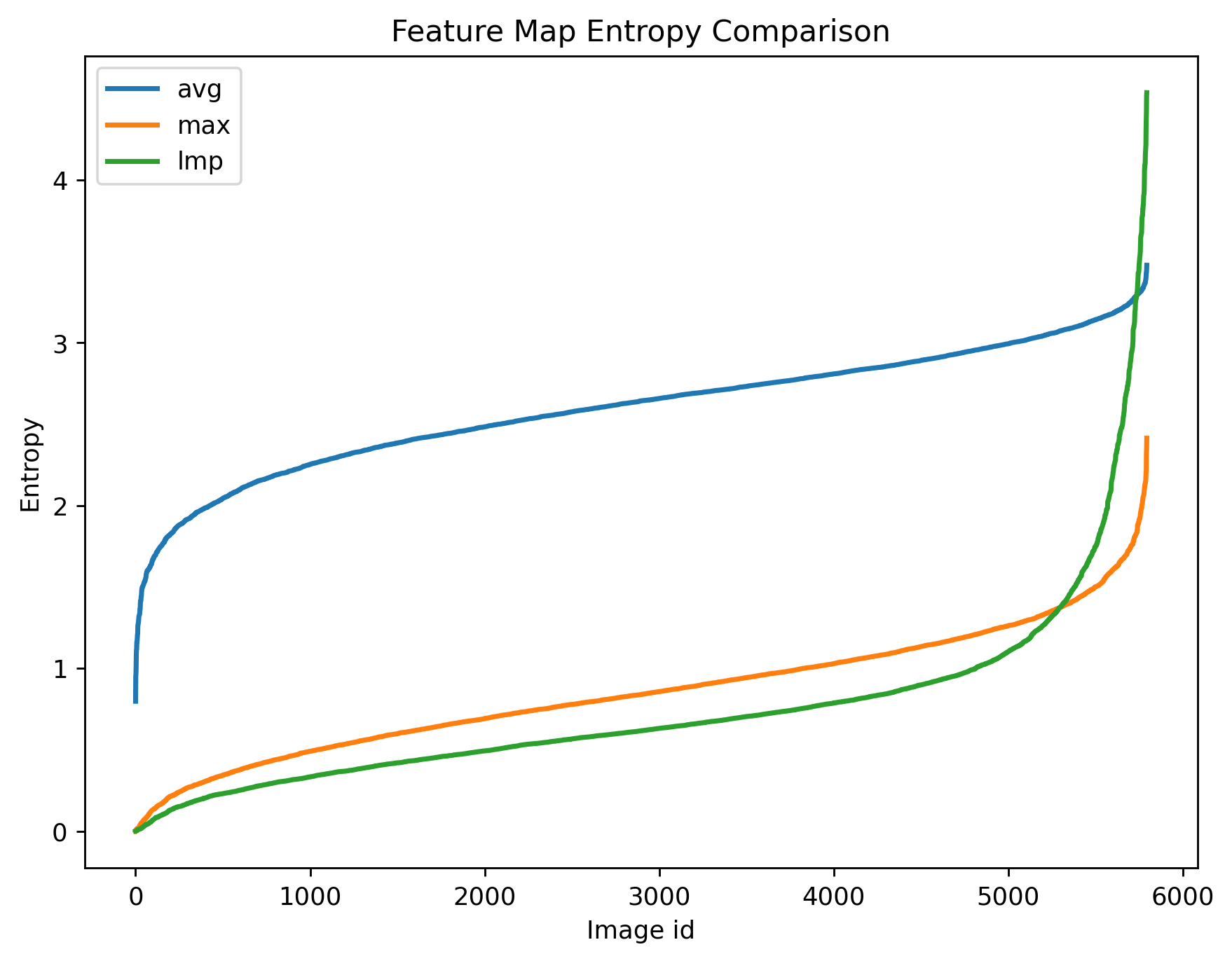}
    \caption{\textbf{Feature map entropy comparison.} Sparser feature maps have smaller entropy.}
    \label{fig:entropy}
\end{figure}

In this section we conduct a thorough analysis of the leaky max pooling layer. We first qualitatively evaluate how leaky max pooling leads to sparse feature maps. Figure~\ref{fig:lmp}~(b) shows typical filter visualization for ResNet-50 models using average, max and leaky max pooling separately. Average pooling learns background and object-level features, max pooling learns part-level features and leaky max pooling learns keypoint level features. A more complete filter visualization can be found in the appendix section. For a quantitative evaluation of the feature map sparsity, we take the normalized final conv-layer feature maps and compute their average entropy: sparser feature maps have smaller entropy. For each pooling method, we show in Figure~\ref{fig:entropy} the average feature map entropy for all images in the test set. It is clear that leaky max pooling $>$ max pooling $>$ average pooling in terms of feature map sparsity.

\begin{table}[h!]
\centering
\begin{tabular}{ | c c c c | } 
 \hline
  AVG & MAX & LMP ($\epsilon=0.1$) & LMP ($\epsilon=0.01$)\\
 \hline
  85.0 & 86.1 & 81.4 & 85.3  \\ 
 \hline
\end{tabular}
\caption{Classification Accuracy for different $\epsilon$}
\label{table:eps}
\end{table}


In leaky max pooling, $\epsilon$ is a hyperparameter  that controls how much penalty is imposed on non-maximum locations. We study how different values of  $\epsilon$ lead to different results. First we compare the classification accuracy using two different $\epsilon$ values 0.1 and 0.01. These results, together with average and max pooling results, are shown in Table~\ref{table:eps}. By visualizing the filters learned by $\epsilon=0.1$ and  $\epsilon=0.01$, we find the when  $\epsilon=0.01$, there are a significantly higher number of filters that learn random features than when $\epsilon=0.1$. Although $\epsilon=0.01$ has a higher accuracy, we choose $\epsilon=0.1$ in all our experiments for its better keypoint detection ability. When $\epsilon$ is larger, leaky max pooling is better at suppressing dense inputs.

\begin{figure*}[ht]
  \centering
  \includegraphics[width=\linewidth]{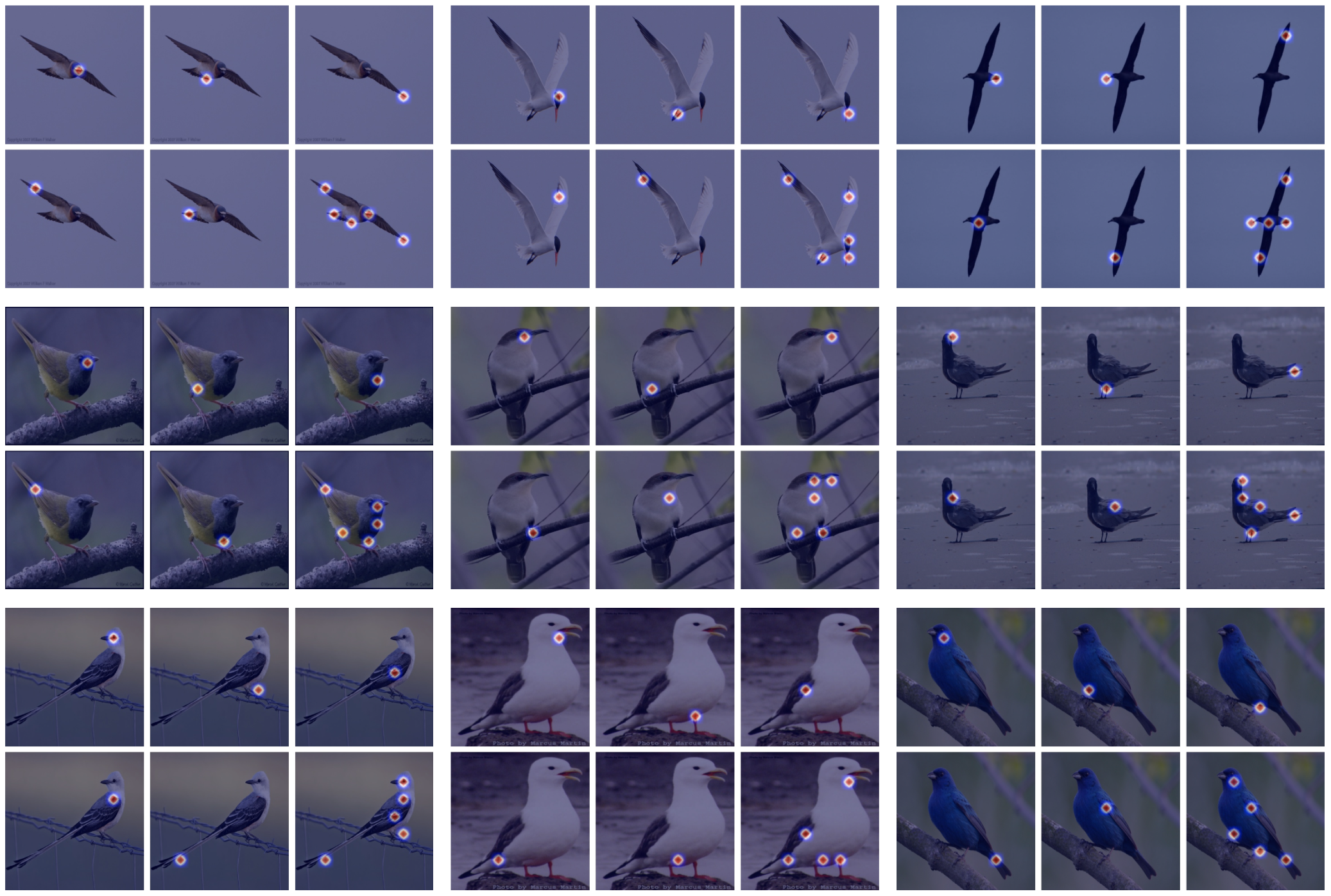}
  \caption{\textbf{LMPNet keypoint prediction (k=5) results on CUB-200-2011.} Note how well LMPNet handles different bird species and poses. More results can be found in the appendix section}
  \label{fig:prediction}
\end{figure*}

\subsection{Keypoint Prediction Evaluation}
In this section, we evaluate how well our keypoint prediction corresponds to ground truth keypoints. The final conv-layer feature maps are small, either $7 \times 7$ or $14 \times 14$, depending on the input size. We set the number of final keypoints to $k=5$ for the purpose of better visualization. We show our final keypoint prediction results in Figure~\ref{fig:prediction}. All of the predicted keypoints are within the object boundary and are well-aligned with head, tail, feet, wing tips, \textit{etc.} Our method handles complex bird pose surprisingly well. Examples can be found in the first row of Figure~\ref{fig:prediction} that shows our keypoint prediction results for flying birds, a setting where other methods typically struggle.

We next consider a quantitative evaluation of the generated keypoint prediction. Inspired by the traditional PCK (percentage of correct keypoint) metric for supervised keypoint prediction, we adopt a greedy matching strategy where a keypoint prediction is regarded as accurate if it is within a small range ($\alpha = 0.1$ of the shorter image side length) of a ground truth keypoint. We call this metric greedy PCK. The erasing threshold in our learnable clustering algorithm determines how many keypoint proposals are erased after one iteration of keypoint prediction. We find that setting a smaller threshold generally leads to higher greedy PCK, but keypoint predictions may overlap with each other when the threshold is too small. We show how the greedy PCK varies for different erasing thresholds in Table~\ref{table:pck}. We choose $thr = 3$ in our final model to balance keypoint duplication and keypoint quality.

\begin{table}[h!]
\centering
\begin{tabular}{ |c | c c c c c || c | } 
 \hline
  & KP 1 & KP 2 & KP 3 & KP 4 & KP 5 & AVG\\
 \hline
 $thr=3$ & 96.2 & 89.2 & 86.6 & 82.4 & 75.4 & 86.0 \\ 
 $thr=2$ & 97.5 & 89.1 & 87.8 & 83.1 & 74.8 & 86.5 \\ 
 $thr=1$ & 96.5 & 91.4 & 91.4 & 90.7 & 87.6 & 91.5 \\ 
 \hline
 \hline
 FCN~\cite{GuoF_WACV2019} & 97.4 & 96.8 & 91.3 & 80.3 & 75.3 & 88.2\\ 
 \hline
\end{tabular}
\caption{\textbf{Greedy PCK @ $\alpha=0.1$ for different erasing threshold.} We choose $thr=3$ to avoid keypoint duplication and maintain high keypoint prediction accuracy. FCN denotes the fully convolutional neural network result~\cite{GuoF_WACV2019} and the 5 keypoints are evenly sampled from their original 15 keypoint predictions}
\label{table:pck}
\end{table}

\section{Conclusion}\label{sec:conclusion}

We propose a novel network architecture named LMPNet for weakly-supervised keypoint discovery using only image class labels as supervision. Novel network components are proposed to explicitly transform discriminatively-trained intermediate network filters into keypoint detectors. Due to the lack of ground truth keypoint annotations, we define semantic object keypoints as ``non-repeatable local patterns'', which leads to a novel leaky max pooling layer derived from a generalized global pooling formulation. The pooling vector contains 1 for the max input location and $-\epsilon$ everywhere else. The proposed leaky max pooling layer encourages the final conv-layer filters to have sparse activations that are well aligned with object keypoints. We show by visualization that discriminatively-trained filters have two issues: consistency and diversity. We first propose a simple yet effective selection strategy to  ensure that filter responses are consistent, and second propose attention mask-out to distribute the network's attention to less-represented regions. A learnable clustering layer is applied to aggregate the keypoint proposals into keypoint predictions, in a sequentially greedy way. The proposed network architecture is shown to produce meaningful object keypoints with only image labels. LMPNet works surprisingly well on complex bird poses and achieves high greedy matching PCK comparable to strongly-supervised baseline models.

{\small
\bibliographystyle{ieee_fullname}
\bibliography{FarrellMendeley,addl_refs}
}
\end{document}